\pdfoutput=1

\documentclass[11pt]{article}

\usepackage[final]{acl}

\usepackage{times}
\usepackage{latexsym}
\usepackage{amsmath}
\usepackage{amssymb}

\DeclareMathOperator*{\argmax}{arg\,max}
\usepackage{enumitem}
\usepackage{array}
\usepackage{booktabs}
\usepackage{multirow}
\usepackage{caption}
\usepackage{makecell}
\usepackage{float}
\usepackage{amsfonts}
\usepackage{hyperref}

\usepackage[dvipsnames]{xcolor}
\usepackage[table]{xcolor}
\usepackage{colortbl}

\definecolor{bluecolor}{RGB}{0,114,178} % for blue text
\definecolor{redcolor}{RGB}{213,94,0}   % for red text
\definecolor{greencolor}{RGB}{0,158,115} % for green text

\usepackage[T1]{fontenc}

\usepackage[utf8]{inputenc}

\usepackage{microtype}

\usepackage{inconsolata}

\usepackage{graphicx}

\title{\makebox[\textwidth][l]{Topic Coverage-based Demonstration Retrieval for In-Context Learning}}

\author{
Wonbin Kweon\(^{1}\) \quad SeongKu Kang\(^{2}\) \quad Runchu Tian\(^{1}\) \quad Pengcheng Jiang\(^{1}\)\\
{\bf Jiawei Han\(^{1}\)} \quad {\bf Hwanjo Yu\(^{3}\)\thanks{Corresponding author}}\\
\(^{1}\)University of Illinois Urbana-Champaign\\
\(^{2}\)Korea University \quad \(^{3}\)Pohang University of Science and Technology \\
\texttt{\{wonbin, runchut2, pj20, hanj\}@illinois.edu}\\
\texttt{seongkukang@korea.ac.kr} \quad \texttt{hwanjoyu@postech.ac.kr}
}

\begin{document}
\maketitle

\begin{abstract}
% The effectiveness of In-context learning (ICL) heavily depends on selecting demonstrations to provide all the required information for the test input.
% However, previous work has limitations in that neither method explicitly examines the required knowledge at a fine-grained level.
% They either select demonstrations based on the embedding similarity or the surface generation probability.
% In this paper, we introduce \textbf{TopicK}, a topical knowledge-aware retrieval framework that selects demonstrations to comprehensively cover the information demands of the test input.
% Specifically, we estimate the required and covered topics using a lightweight topic predictor, without relying on human labels or LLM inference at test time.
% Building on these components, we devise a novel relevance score, that is equivalent to minimizing model uncertainty for the test input.

The effectiveness of in-context learning relies heavily on selecting demonstrations that provide all the necessary information for a given test input.
To achieve this, it is crucial to identify and cover fine-grained knowledge requirements. 
However, prior methods often retrieve demonstrations based solely on embedding similarity or generation probability, resulting in irrelevant or redundant examples.
In this paper, we propose \textbf{TopicK}, a topic coverage-based retrieval framework that selects demonstrations to comprehensively cover topic-level knowledge relevant to both the test input and the model.
Specifically, TopicK estimates the topics required by the input and assesses the model’s knowledge on those topics.
TopicK then iteratively selects demonstrations that introduce previously uncovered required topics, in which the model exhibits low topical knowledge.
We validate the effectiveness of TopicK through extensive experiments across various datasets and both open- and closed-source LLMs.
Our source code is available at \textcolor{blue}{\url{https://github.com/WonbinKweon/TopicK_EMNLP2025}}
\end{abstract}

\section{Introduction}
%% 1. Demonstration selection in important for ICL
Large language models (LLMs) \citep{llama3,qwen2.5,gpt4o} have demonstrated a remarkable capacity to internalize and utilize novel information solely from contextual input, without requiring any parameter updates.
This ability, referred to as \textit{in-context learning} (ICL) \cite{few-shot}, enables LLMs to leverage a small set of input-output demonstrations to solve previously unseen tasks or adapt to new domains.
However, prior studies \citep{liu-etal-2022-makes, peng-etal-2024-revisiting} have shown that the effectiveness of ICL is highly sensitive to the choice of these demonstrations.
Consequently, identifying the most informative demonstrations is critical to fully realizing the generalization potential of LLMs through ICL.

%% 2. Early: data-aware approaches 
% In early work, \textit{similarity-based} approaches employed retrieval modules to select demonstrations from a candidate pool that are relevant to the given test input.
% \citet{bm25} utilized the BM25 retriever to select exemplars with high lexical overlap, and \citet{liu-etal-2022-makes} adopted a dense retriever to identify $K$-nearest-neighbors in the embedding space.
% These approaches encode the test input and candidate demonstrations separately (i.e., \textit{dual-encoder}), enabling efficient retrieval with low latency.
% However, such off-the-shelf retrievers operate independently of the inference models (i.e., LLMs), thus failing to account for their parametric knowledge.
In early work, \textit{similarity-based} approaches \citep{gao-etal-2021-making, liu-etal-2022-makes, ceil, gupta-etal-2023-coverage} employed retrieval modules to select relevant demonstrations from a candidate pool, given a test input.
They either utilize a BM25 retriever \citep{bm25} to select exemplars with high lexical overlap, or dense retrievers \citep{sbert, roberta} to identify $K$-nearest-neighbors in the embedding space.
These approaches encode the test input and candidate demonstrations separately, enabling efficient retrieval with low latency. % (i.e., \textit{dual-encoder})
However, such off-the-shelf retrievers operate independently of the inference models (i.e., LLMs), thus failing to account for their parametric knowledge.

% %% 3. Recent: model-aware approaches
% To address this limitation, recent \textit{model-aware} approaches~\citep{iter-etal-2023-context, wang-etal-2024-mdr, peng-etal-2024-revisiting} propose selecting demonstrations that reduce the LLM’s predictive uncertainty.
% They measure the generation probability of either the test input~\citep{peng-etal-2024-revisiting} or the model output~\citep{iter-etal-2023-context}, conditioned on each candidate.
% While this aligns retrieval with LLMs, it requires a separate inference for every test–candidate pair (i.e., \textit{cross-encoder}), incurring a substantial computational burden at test time.
% Moreover, by focusing only on instance-level informativeness, they fail to capture fine-grained aspects (e.g., topics) within each instance.
%% 3. Recent: uncertainty-based approaches
To address this limitation, recent \textit{uncertainty-based} approaches~\citep{iter-etal-2023-context, wang-etal-2024-mdr, peng-etal-2024-revisiting} propose selecting demonstrations that reduce the LLM’s predictive uncertainty.
They measure the generation probability of either the test input~\citep{peng-etal-2024-revisiting} or the model output~\citep{iter-etal-2023-context}, conditioned on each candidate.
Demonstrations are then ranked based on these probabilities.
While this aligns retrieval with LLMs, it requires a separate LLM inference for every test-candidate pair, incurring a substantial computational burden at test time.
Moreover, as demonstrations are ranked by independently computed probabilities, these methods fail to ensure diversity among the selected demonstrations.

\begin{figure*}[t]
  \centering
  \includegraphics[width=0.95\linewidth]{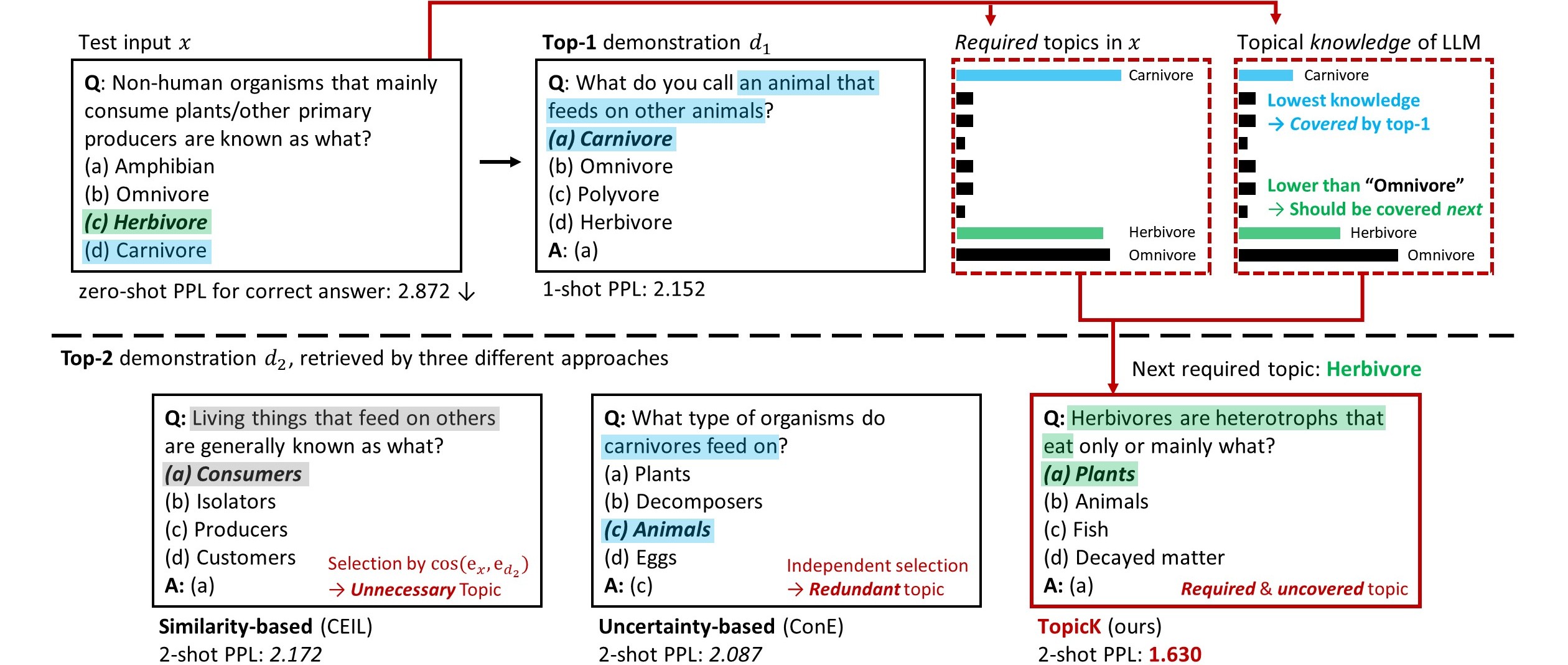}
  \caption {Case study on SciQ dataset and Llama-3.2-1B. The top-1 demonstration is given the same for all three methods. PPL denotes the perplexity (lower is better) for the correct answer \texttt{`(c) Herbivore'}.}
\label{fig:intro}
\end{figure*}

%% 4. Motivational case study
Figure~\ref{fig:intro} presents a motivating case study for a test input from SciQ~\citep{SciQ}.
The similarity-based approach~\citep{ceil} retrieves a demonstration solely based on embedding similarity, thereby failing to capture the specific topics required by the test input.
Meanwhile, the uncertainty-based method~\citep{peng-etal-2024-revisiting} selects a redundant demonstration about \texttt{`Carnivore'}, where the model exhibits the highest uncertainty, overlooking the diversity among demonstrations.
These shortcomings highlight the need for a novel approach that identifies fine-grained knowledge requirements (e.g., topics), and thoroughly covers them by retrieving relevant yet diverse demonstrations.

%% 5. Propose: TopicK
We propose \textbf{TopicK}, a topic coverage-based demonstration retrieval framework that explicitly captures the fine-grained informational demands of both the test input and the model.
Specifically, TopicK estimates three key components:
\noindent (1) \textit{required topics} in the test input,
\noindent (2) \textit{covered topics} in candidate demonstrations, and
\noindent (3) \textit{topical knowledge} encoded in the model.
These components are inferred via a lightweight topic predictor, without requiring human annotations or LLM inference at test time.
TopicK then iteratively selects demonstrations that introduce previously uncovered required topics, for which the model shows low topical knowledge.
As a result, in Figure~\ref{fig:intro}, TopicK achieves the lowest perplexity by retrieving a demonstration with a new topic \texttt{`Herbivore'}.

%% 6. Bullets
The key features of TopicK are summarized as:
\begin{itemize}[leftmargin=10pt, itemsep=0pt]
% \item TopicK captures fine-grained topic-level knowledge requirements of test inputs, going beyond existing instance-level informativeness measures.
\item TopicK captures fine-grained topic-level knowledge requirements of test inputs, going beyond existing methods that rely solely on embedding similarity or generation probability.
\item TopicK infers the required topics using a lightweight topic predictor, avoiding the need for LLM inference at test time as in previous uncertainty-based methods.
\item TopicK consistently outperforms state-of-the-art approaches across diverse benchmarks and model scales, including both open- and closed-source LLMs.
\end{itemize}

\section{Preliminary}
\subsection{In-Context Learning}
In-context learning (ICL) is one of the core emergent capabilities of large language models (LLMs), enabling them to internalize and utilize novel information solely from contextual input, without requiring updates to model parameters.

\paragraph{Problem Formulation} 
Given a test input $x$, an LLM generates the output $\hat{y}$, conditioned on a few in-context demonstrations, as follows:
\begin{equation}
    \hat{y} \sim p_{\text{LM}}(\hat{y} \mid \underbrace{d_1, d_2, \ldots, d_K}_{\text{demonstrations}}, x),
\end{equation}
where each demonstration $d_i = (x_i, y_i)$ is selected from a candidate pool $\mathcal{D} = \{(x_i, y_i)\}_{i=1}^N$.

\vspace{+0.2cm}
\noindent A variety of strategies have been proposed to improve ICL performance, including demonstration selection~\citep{gao-etal-2021-making, liu-etal-2022-makes, rubin-etal-2022-learning, wu-etal-2023-self}, demonstration ordering~\citep{lu-etal-2022-fantastically, lee-etal-2024-crafting, pearl}, and prompt template design~\citep{deng-etal-2022-rlprompt, xu-etal-2022-zeroprompt, prasad-etal-2023-grips,cheng-etal-2023-uprise}.
In this work, we focus on demonstration selection, which has been identified as the most critical factor influencing ICL effectiveness~\citep{peng-etal-2024-revisiting, wan2024teach}.

\subsection{Similarity-based Approaches}
% Early work leverages retrieval modules to select demonstrations that are semantically similar to the test input.
% \citet{gao-etal-2021-making} and \citet{liu-etal-2022-makes} employ an embedding model \citep{sbert} to obtain vector representations $\mathbf{e}_x$ and $\mathbf{e}_d$ for the test input $x$ and each demonstration $d \in \mathcal{D}$.
Early work \citep{gao-etal-2021-making, liu-etal-2022-makes} employs embedding models \citep{sbert, roberta} to obtain vector representations $\mathbf{e}_x$ and $\mathbf{e}_d$ for the test input $x$ and each demonstration $d \in \mathcal{D}$.
The top-$K$ demonstrations are then selected by ranking candidates in descending order of cosine similarity, i.e., $\cos\left( \mathbf{e}_x, \mathbf{e}_d \right)$.
Subsequent studies further take account of diversity by employing majority voting \citep{hongjin2022selective}, determinantal point processes \citep{ceil}, or BERTScore \cite{gupta-etal-2023-coverage}. %  to enhance diversity among selected exemplars.

% \jh{The citation entry: hongjin2022selective should be fixed, the author list should be:
% Hongjin Su, Jungo Kasai, Chen Henry Wu, Weijia Shi, Tianlu Wang, Jiayi Xin, Rui Zhang, Mari Ostendorf, Luke Zettlemoyer, Noah A. Smith, Tao Yu, with the same title: Selective Annotation Makes Language Models Better Few-Shot Learners.  It should be Su et al.
% }

\paragraph{Limitation}
While these approaches offer fast retrieval, they remain model-independent and overlook the parametric knowledge of LLMs.
Moreover, they consider diversity only at the surface-level embedding space.
As a result, although the retrieved demonstrations may be semantically similar to the test input, they often provide limited utility and fail to meaningfully influence the model’s decision-making process \citep{peng-etal-2024-revisiting}.

% While these dual-encoder approaches offer fast retrieval and ease of implementation, they remain model-independent and fail to account for the LLM’s internal reasoning process.
% Consequently, even demonstrations semantically close to the test input may offer limited utility if they do not meaningfully contribute to the model's decision-making process \citep{peng-etal-2024-revisiting}.

\begin{figure*}[ht!]
  \centering
  \includegraphics[width=1\linewidth]{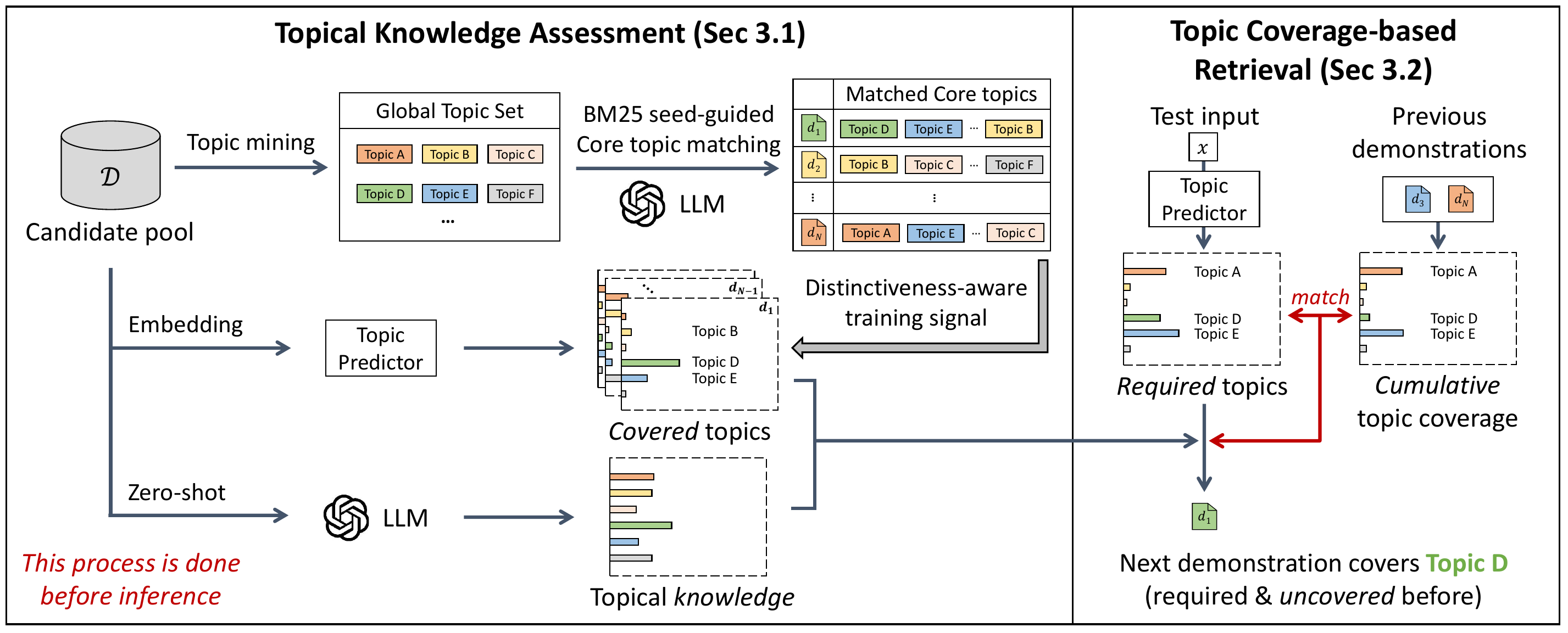}
  \caption {Overview of topic coverage-based demonstration retrieval (TopicK) framework.}
  \vspace{-0.3cm}
\label{fig:method}
\end{figure*}

\subsection{Uncertainty-based Approaches}
Recent approaches argue that the utility of a demonstration is not solely determined by its similarity to the test input, but also by how it interacts with LLMs \citep{peng-etal-2024-revisiting, chen-etal-2024-learning-retrieve}.
These uncertainty-based approaches aim to select demonstrations that explicitly reduce the model’s predictive uncertainty.
For instance, \citet{iter-etal-2023-context} select demonstrations that minimize the entropy of the model's output distribution: $\arg\min_{d_i \in \mathcal{D}} \; \mathbb{H}(\hat{y} \mid d_i, x)$. %, conditioned on both the test input and \textit{each} candidate demonstration
Similarly, \citet{peng-etal-2024-revisiting} select candidates that minimize the entropy of the test input: $\arg\min_{d_i \in \mathcal{D}} \; \mathbb{H}(x \mid d_i)$. %assess the informativeness of each demonstration by its ability to explain the test input itself, 

\paragraph{Limitation}
% Although uncertainty-based objectives better align retrieval with the model’s generative behavior, they require a separate LLM inference to examine each candidate demonstration.
Although uncertainty-based objectives align demonstration retrieval with LLMs, they require a separate LLM inference to examine each demonstration.
This incurs substantial computational overhead, severely limiting scalability and practical deployment.
Furthermore, they rank the demonstrations based on independently computed probabilities, overlooking the diversity in the selected demonstrations.

\section{Methodology}
We propose \textbf{TopicK}, a novel demonstration retrieval framework that leverages topics to explicitly examine the fine-grained informational demands of both the test input and the target LLM.
TopicK consists of two major stages as follows:
\begin{itemize}[leftmargin=10pt, itemsep=0pt]
% \item \textbf{Topical Knowledge Assessment} (\S\ref{sec3.1}): we introduce a framework to estimate three key components (1) \textit{required topics} in the test input, (2) \textit{covered topics} in demonstrations, and (3) \textit{topical knowledge} encoded in the LLM’s parameters.
\item \textbf{Topical Knowledge Assessment} (\S\ref{sec3.1}): TopicK estimates three key components (1) \textit{required topics} in the test input, (2) \textit{covered topics} in candidate demonstrations, and (3) \textit{topical knowledge} encoded in the LLM’s parameters.
% \item \textbf{Topic Coverage-based Retrieval} (\S\ref{sec3.2}): we describe a method for selecting demonstrations that introduce previously uncovered required topics, where the model exhibits low topical knowledge.
\item \textbf{Topic Coverage-based Retrieval} (\S\ref{sec3.2}): TopicK selects demonstrations that introduce previously uncovered required topics, where the model exhibits low topical knowledge.
\end{itemize}

\subsection{Topical Knowledge Assessment}\label{sec3.1}
We first identify core topics of each demonstration, without relying on external data or human annotations.
Then, a lightweight topic predictor is devised based on the identified topics, and utilized to estimate the topic distributions of both the test input and candidate demonstrations.
% Lastly, we examine the required topics, covered topics, and topical knowledge, by utilizing the topic predictor.

\subsubsection{Topic Identification}\label{sec3.1.1}
Given a candidate pool $\mathcal{D}$ and a topic set $\mathcal{T}$, our objective is to identify the core topics of each demonstration.
While any pre-defined topic set \citep{shen-etal-2018-web} can be employed, for broader applicability, we construct $\mathcal{T}$ from scratch using topic mining tools \citep{autophrase, seedtopicmine}.

\paragraph{Candidate Topic Matching}
After constructing $\mathcal{T}$, we find a candidate topic set for each demonstration $d \in \mathcal{D}$ with two types of matching:
\begin{itemize}[leftmargin=10pt, itemsep=0pt]
    \item \textbf{Lexical Overlap:} Select the top-10 topics based on BM25 scores~\citep{bm25}.
    \item \textbf{Semantic Similarity:} Select the top-10 topics based on cosine similarity $\cos(\mathbf{e}_d, \mathbf{e}_t)$. ($t \in \mathcal{T}$)
\end{itemize}
To ensure coverage, topics matched by the lexical overlap are excluded from the semantic similarity matching.
The candidate topic set $\mathcal{T}'_d \subset \mathcal{T}$ is obtained by merging those two results.

\paragraph{Core Topic Matching with LLMs}\label{sec:coretopic}
Topics can exhibit varying semantics depending on the context.
To consider this, we leverage the contextualization capabilities of LLMs.
Specifically, we \hyperref[tab:prompt]{prompt} GPT-4o~\citep{gpt4o} to select the core topics from $\mathcal{T}'_d$ and identify any missing but relevant topics. %% prompt in appendix
This process yields the finalized core topic set $\mathcal{T}_d \subset \mathcal{T}$ for each demonstration $d \in \mathcal{D}$.

\subsubsection{Topic Predictor}\label{sec3.1.2}
Using the identified core topics, we devise a lightweight topic predictor that maps each demonstration embedding $\mathbf{e}_d$ to a topic distribution $\hat{\mathbf{t}}_d \in [0,1]^{|\mathcal{T}|}$.
% Each element $\hat{\mathbf{t}}_{d,t} \in \hat{\mathbf{t}}_d$ represents the degree of membership of topic $t$ in the topic set $\mathcal{T}_d$.
Each element $\hat{\mathbf{t}}_{d,t} \in \hat{\mathbf{t}}_d$ represents the degree of membership of topic $t$ in the core topic set $\mathcal{T}_d$.
In this work, we employ a three-layer MLP $\hat{\mathbf{t}}_d = f(\mathbf{e}_d)$, as the simplest choice. 
% We note that the topic predictor not only generalizes to unseen test inputs but also \textit{enriches} topic distributions by inferring related topics beyond the initial core set, capturing semantic similarities between them.
We note that the topic predictor not only generalizes to unseen test inputs (i.e., $\mathbf{e}_x$), but also \textit{enriches} the topic distributions of candidates by inferring related topics beyond the initial core topic set. %, capturing semantic similarities between them.

\paragraph{Distinctiveness-aware Training Signal}
A naive training signal for $\hat{\mathbf{t}}_{d}$ would be a binary vector $\mathbf{t}_d \in \{0,1\}^{|\mathcal{T}|}$, where $\mathbf{t}_{d,t} = 1$ if $t \in \mathcal{T}_d$ and $0$ otherwise.
However, not all topics contribute equally; some topics are more distinctive to a given demonstration.
To capture this, we adopt a distinctiveness metric inspired by \citet{taxocom}:
\begin{equation}
\text{DST}(d,t) = \frac{\exp\left( \text{BM25}(d,t) \right)}{1 + \sum_{d' \in \mathcal{D}_d} \exp\left( \text{BM25}(d',t) \right)},
\end{equation}
where $\mathcal{D}_d$ denotes the set of 100 demonstrations nearest to $d$ in the embedding space.
We then normalize the distinctiveness scores to produce a soft target vector $\mathbf{t}_d \in [0,1]^{|\mathcal{T}|}$: %via min-max scaling 
\begin{equation}
\mathbf{t}_{d,t} = \frac{\text{DST}(d,t)}{\max_{t' \in \mathcal{T}_d} \text{DST}(d,t') }.
\label{eq:softlabel}
\end{equation}
Finally, we train the topic predictor $\hat{\mathbf{t}}_d = f(\mathbf{e}_d)$ by using a binary cross-entropy loss:
% \begin{equation}
% \mathcal{L}_{\text{TP}} = - \sum_{d \in \mathcal{D}} \sum_{t \in \mathcal{T}} \mathbf{t}_{d,t} \log \hat{\mathbf{t}}_{d,t} + (1-\mathbf{t}_{d,t}) \log (1-\hat{\mathbf{t}}_{d,t}).
% \end{equation}
\begin{equation}
\mathcal{L}_{\text{TP}} = - \sum_{d \in \mathcal{D}} \Big( \sum_{t \in \mathcal{T}_d} \mathbf{t}_{d,t} \log \hat{\mathbf{t}}_{d,t} +  \sum_{t \notin \mathcal{T}_d} \log (1-\hat{\mathbf{t}}_{d,t})  \Big).
\label{eq:BCE}
\end{equation}

\subsubsection{Topical Knowledge Assessment}
\paragraph{Required \& Covered Topics}
To assess relevant topics for each sample, we utilize the trained topic predictor described earlier.
Given embeddings of a test input $\mathbf{e}_{x}$ and a demonstration $\mathbf{e}_d$, we predict their topic distributions $\hat{\mathbf{t}}_x = f(\mathbf{e}_x) \in [0,1]^{|\mathcal{T}|}$ and $\hat{\textbf{t}}_d = f(\mathbf{e}_d) \in [0,1]^{|\mathcal{T}|}$.
Here, $\hat{\mathbf{t}}_x$ represents the \textit{required} topics needed to understand and answer the test input $x$, while $\hat{\mathbf{t}}_d$ indicates the \textit{covered} topics in the candidate demonstration $d$.
These distributions allow fine-grained assessment of how well a demonstration aligns with the informational needs of a test input.

\paragraph{Topical Knowledge}
In addition to the required and covered topics, we also consider the model's inherent knowledge on each topic, defined as $\hat{\mathbf{t}}_{\text{LM}} \in [0,1]^{|\mathcal{T}|}$.
%% while there have been ~~. simple yet effective direct.
We estimate the topical knowledge by aggregating the model’s \textit{zero-shot} accuracy on candidate demonstrations:
\begin{equation}
\begin{aligned}
       \hat{\mathbf{t}}_{\text{LM}, t}  & = \frac{\sum_{d \in \mathcal{D}} \hat{\mathbf{t}}_{d,t} \cdot \text{zero-shot}(d)}{\sum_{d \in \mathcal{D}} \hat{\mathbf{t}}_{d,t}}, \\
       \text{zero-shot}(d) & = \mathbf{1}[y = \argmax_{\hat{y}}p_{\text{LM}}(\hat{y}|x)],
\end{aligned}
\end{equation}
where $\text{zero-shot}(d) \in \{0,1\}$ indicates the zero-shot accuracy on demonstration $d = (x,y) \in \mathcal{D}$.
That is, we measure how reliably the LLM answers instances associated with each topic without any demonstrations.
This prior provides insights into which topics the model has already internalized, allowing us to avoid selecting demonstrations for topics that the model already knows well.

% \subsection{Topical Knowledge-Aware Retrieval}\label{sec3.2}
\subsection{Topic Coverage-based Retrieval}\label{sec3.2}
\subsubsection{Topic Coverage-aware Relevance}\label{sec3.2.1}
We define a novel relevance score between a test input $x$ and a candidate demonstration $d$ as follows:
% \begin{equation}
%     r(x,d) = \sum_{t \in \mathcal{T}} \frac{\hat{\mathbf{t}}_{x,t} \cdot \hat{\mathbf{t}}_{d,t}}{\hat{\mathbf{t}}_{\text{LM}, t}}.
% \label{eq:relevance}
% \end{equation}
\begin{equation}
    % r(x,d) = \sum_{t \in \mathcal{T}} \frac{\hat{\mathbf{t}}_{x,t} \cdot \hat{\mathbf{t}}_{d,t}}{\hat{\mathbf{t}}_{\text{LM}, t}} = (\hat{\mathbf{t}}_{x} \odot \hat{\mathbf{t}}_{d}) \oslash \hat{\mathbf{t}}_{\text{LM}},
    r(x,d) = \sum_{t \in \mathcal{T}} \frac{\hat{\mathbf{t}}_{x,t} \cdot \hat{\mathbf{t}}_{d,t}}{\hat{\mathbf{t}}_{\text{LM}, t}} = \langle \hat{\mathbf{t}}_{x} \oslash \hat{\mathbf{t}}_{\text{LM}}, \hat{\mathbf{t}}_{d} \rangle,
\label{eq:relevance}
\end{equation}
\noindent where $\oslash$ denotes the element-wise division and $\langle\cdot,\cdot\rangle$ is the inner product.
This relevance score captures three critical aspects:
\begin{itemize}[leftmargin=10pt, itemsep=0pt]
    \item \textbf{Required Topics}: $\hat{\mathbf{t}}_{x,t}$ prioritizes topics highly relevant to the test input.
    \item \textbf{Covered Topics}: $\hat{\mathbf{t}}_{d,t}$ promotes demonstrations that provide high coverage of required topics.
    \item \textbf{Topical Knowledge}: $\hat{\mathbf{t}}_{\text{LM},t}$ down-weights topics that the model already knows well.
\end{itemize}
% \noindent By design, TopicK selects a demonstration $d$ whose covered topics $\hat{\mathbf{t}}_d$ align well with those of the test input $\hat{\mathbf{t}}_x$.
\noindent By our design, TopicK assigns a high relevance score for a demonstration $d$, whose covered topics $\hat{\mathbf{t}}_d$ align well with the knowledge-weighted required topics (i.e., $\hat{\mathbf{t}}_{x} \oslash \hat{\mathbf{t}}_{\text{LM}}$).
% It is worth noting that $\hat{\mathbf{t}}_d$ and $\hat{\mathbf{t}}_{\text{LM}}$ are pre-computed, and only $\hat{\mathbf{t}}_x$ is inferred via a lightweight topic predictor for the retrieval.
It is worth noting that $\hat{\mathbf{t}}_{\text{LM}}$ is pre-computed before the test time, while $\hat{\mathbf{t}}_x$ and $\hat{\mathbf{t}}_d$ are inferred via a lightweight topic predictor.
Therefore, \textbf{TopicK enables LLM-aware demonstration selection without LLM inference at test time.}
We use the final relevance score as $r(x,d) + \lambda \cdot \cos(\mathbf{e}_x, \mathbf{e}_d)$, incorporating both topical and semantic relevance.\label{eq:lambda}

\subsubsection{Cumulative Topic Coverage}\label{sec3.2.2}
We further incorporate cumulative topic coverage to avoid retrieving redundant demonstrations.
% Given a set of previously selected demonstrations $\mathcal{D}'_x$, we update the relevance score from Eq.~\ref{eq:relevance} as:
% \begin{equation}
% \begin{aligned}
%     r(x,d \mid \mathcal{D}'_x) = \langle \hat{\mathbf{t}}_{x} \oslash \hat{\mathbf{t}}_{\text{LM}}, \,\hat{\mathbf{t}}_{d \cup  \mathcal{D}'_x} \ominus \hat{\mathbf{t}}_{\mathcal{D}'_x} \rangle.
% \end{aligned}
% \label{eq:diversity}
% \end{equation}
Given a set of previously selected demonstrations $\mathcal{D}'_x$, we update the covered topics $\hat{\mathbf{t}}_{d}$ in Eq.~\ref{eq:relevance} as:
\begin{equation}
\begin{aligned}
    \hat{\mathbf{t}}_{d} \leftarrow (\hat{\mathbf{t}}_{d \cup  \mathcal{D}'_x} -\hat{\mathbf{t}}_{\mathcal{D}'_x}).
\end{aligned}
\label{eq:diversity}
\end{equation}
\noindent Here, $\hat{\mathbf{t}}_{\mathcal{D}'_x}$ and $\hat{\mathbf{t}}_{d \cup \mathcal{D}'_x}$ represent the cumulative topic coverage before and after adding $d$.
These are also obtained by the topic predictor using mean-pooled embeddings, e.g., $\hat{\mathbf{t}}_{d \cup \mathcal{D}'_x} = f(\mathbf{e}_{d \cup \mathcal{D}'_x})$ and $\mathbf{e}_{d \cup \mathcal{D}'_x} = (\mathbf{e}_d + \sum_{d' \in \mathcal{D}'_x} \mathbf{e}_{d'}) / (1 + |\mathcal{D}'_x|)$.
This formulation encourages the selection of the next demonstration that introduces novel topic coverage beyond what has already been covered by $\mathcal{D}'_x$.
After iteratively selecting $K$ demonstrations, the final set $\mathcal{D}_x = \{d_i\}_{i=1}^K$ is prepended to the test input to generate the output $\hat{y} \sim p_{\text{LM}}(\hat{y} \mid \mathcal{D}_x, x)$.
To reduce computational overhead, we retain only the top-300 candidates of the first iteration.

\subsubsection{Theoretical Justification}
Lastly, we provide a theoretical justification for how our topic coverage-aware relevance is derived.
We start from $\mathbb{H}(x|d)$, the uncertainty regarding the test input $x$ given the demonstration $d$.
Since $x$ is known at test time, minimizing this uncertainty is equivalent to maximizing the generation probability $p(x|d)$.
While ConE~\citep{peng-etal-2024-revisiting} estimates this probability through expensive LLM inference at test time, we instead decompose it via topic modeling~\citep{LDA, kang2025improving}:
\begin{equation}
\begin{aligned}
p(x|d) &= \sum_{t \in \mathcal{T}} \, p(x|t) \cdot p(t|d) \\
&= \sum_{t \in \mathcal{T}} \left( p(t|x) \, \cdot \,  p(x) \, / \, p(t) \right) \cdot p(t|d) \\
&= p(x) \cdot \sum_{t \in \mathcal{T}} \underbrace{p(t|x)}_{\substack{\text{required} \\ \text{topics}}} \cdot \underbrace{p(t|d)}_{\substack{\text{covered} \\ \text{topics}}} \, / \, \underbrace{p(t)}_{\substack{\text{topical} \\ \text{knowledge}}}.
\end{aligned}
\end{equation}
\noindent Here, $p(x)$ is constant across demonstrations.
The terms $p(t|x)$, $p(t|d)$, and $p(t)$ correspond to $\hat{\mathbf{t}}_{x,t}$, $\hat{\mathbf{t}}_{d,t}$, and $\hat{\mathbf{t}}_{\text{LM},t}$ in Eq.~\ref{eq:relevance}, respectively.
Thus, our topic coverage-based retrieval is equivalent to minimizing the model's uncertainty on the test input.

\section{Experiment}
\subsection{Experimental Setup}
Due to a lack of space, please refer to Appendix~\ref{sec:expdetail} for further details.

\paragraph{Models}
We conduct experiments using two widely adopted model families, \textbf{Llama3.2} \citep{llama3} and \textbf{Qwen2.5} \citep{qwen2.5}, covering a range of model sizes: Llama-3.2-1B, Llama-3.2-3B, Llama-3.1-8B, Qwen-2.5-0.5B, Qwen-2.5-3B, and Qwen-2.5-7B.
All models are instruction-tuned.
Additionally, we adopt \textbf{Gemini-2.0-Flash-Lite} \citep{gemini}, \textbf{Claude-3.0-Haiku} \citep{claude3}, and \textbf{GPT-4o-mini} \citep{gpt4o} for evaluation on closed-source LLMs.

\begin{table*}[t]
\centering
\resizebox{\textwidth}{!}{
\begin{tabular}{c|ccc|ccc|ccc|ccc|ccc|ccc}
\toprule
& \multicolumn{3}{c|}{\textbf{CommonsenseQA}} 
& \multicolumn{3}{c|}{\textbf{SciQ}} 
& \multicolumn{3}{c|}{\textbf{QNLI}} 
& \multicolumn{3}{c|}{\textbf{MNLI}} 
& \multicolumn{3}{c|}{\textbf{MedMCQA}} 
& \multicolumn{3}{c}{\textbf{Law}} \\
\toprule
\toprule
\textit{Llama3.2} & 1B & 3B & 8B & 1B & 3B & 8B & 1B & 3B & 8B & 1B & 3B & 8B & 1B & 3B & 8B & 1B & 3B & 8B \\
\midrule
Zero & 37.67 & 55.13 & 64.70 & 64.10 & 81.50 & 91.90 & 51.11 & 51.73 & 53.73 & 42.14 & 43.45 & 44.11 & 37.71 & 52.45 & 58.26 & 41.55 & 79.50 & 87.40 \\
Rand & 41.03 & 56.59 & 63.47 & 64.90 & 82.00 & 92.00 & 53.54 & 60.34 & 72.82 & 37.87 & 42.21 & 46.42 & 35.76 & 52.31 & 59.23 & 42.70 & 90.70 & 93.60 \\
BM25 & 42.37 & 52.99 & 64.05 & 67.20 & 82.90 & 92.30 & 55.27 & 68.15 & 75.68 & 41.80 & 46.54 & 50.79 & 38.64 & 56.36 & 61.57 & 44.10 & 91.20 & 94.70 \\
TopK & 43.14 & 56.33 & 65.59 & 71.20 & 89.00 & 92.90 & 60.18 & 71.05 & 77.95 & 50.58 & 58.94 & 66.25 & 39.80 & 59.65 & 67.89 & 47.00 & 91.80 & 96.30 \\
CEIL & 44.18 & 57.78 & 66.68 & 72.20 & 89.20 & 93.30 & 61.06 & 71.46 & 78.63 & 51.22 & 60.04 & 67.02 & 40.09 & 61.25 & 68.10 & 48.10 & 92.25 & 96.80 \\ %%
Set-BSR & 44.72 & 58.48 & 67.49 & 72.90 & 90.20 & 94.40 & 61.80 & 72.32 & 79.59 & 51.84 & 60.77 & 67.84 & 40.27 & 61.79 & 68.93 & 48.70 & 93.00 & 97.35 \\ %%
MDL & 44.51 & 58.23 & 67.10 & 72.60 & 89.80 & 94.30 & 61.34 & 72.17 & 79.81 & 51.76 & 60.34 & 67.97 & 40.16 & 61.51 & 68.79 & 48.50 & 93.10 & 97.25 \\
MDR & 44.46 & 57.78 & 66.88 & 72.40 & 89.60 & 94.10 & 61.22 & 72.14 & 79.37 & 51.66 & 60.27 & 67.88 & 40.25 & 60.88 & 68.63 & 48.35 & 92.85 & 97.10 \\
ConE & 44.34 & 58.40 & 66.91 & 72.80 & 90.10 & 94.50 & 61.56 & 72.20 & 80.14 & 51.89 & 60.53 & \textbf{68.11} & 40.45 & 62.07 & 69.03 & 48.60 & 93.15 & 97.45 \\
TopicK & \textbf{46.19}$^*$ & \textbf{60.52}$^*$ & \textbf{68.63}$^*$ & \textbf{74.60}$^*$ & \textbf{91.20}$^*$ & \textbf{95.20}$^*$ & \textbf{62.51}$^*$ & \textbf{73.55}$^*$ & \textbf{81.35}$^*$ & \textbf{52.81}$^*$ & \textbf{61.67}$^*$ & 68.06 & \textbf{41.80}$^*$ & \textbf{62.36}$^+$ & \textbf{70.21}$^*$ & \textbf{51.70}$^*$ & \textbf{93.80}$^+$ & \textbf{97.60} \\
\toprule
\toprule
\textit{Qwen2.5} & 0.5B & 3B & 7B & 0.5B & 3B & 7B & 0.5B & 3B & 7B & 0.5B & 3B & 7B & 0.5B & 3B & 7B & 0.5B & 3B & 7B \\
\midrule
Zero & 43.41 & 63.44 & 69.45 & 65.10 & 93.00 & 93.80 & 57.97 & 64.31 & 54.15 & 47.20 & 47.78 & 49.54 & 34.16 & 51.24 & 55.47 & 41.10 & 69.70 & 81.85 \\
Rand & 44.72 & 64.50 & 69.21 & 71.00 & 92.60 & 94.60 & 55.39 & 70.52 & 65.41 & 48.61 & 48.14 & 50.02 & 35.19 & 51.28 & 59.59 & 42.00 & 86.90 & 92.10 \\
BM25 & 45.62 & 66.49 & 70.35 & 72.30 & 93.00 & 94.90 & 58.13 & 72.96 & 74.18 & 50.55 & 62.46 & 64.39 & 37.07 & 51.71 & 60.83 & 45.50 & 93.40 & 95.20 \\
TopK & 48.14 & 65.23 & 70.42 & 78.30 & 93.30 & 95.10 & 59.02 & 76.36 & 79.57 & 51.59 & 67.61 & 73.55 & 38.70 & 53.54 & 62.88 & 46.90 & 95.70 & 96.30 \\
CEIL & 49.64 & 66.39 & 71.28 & 80.50 & 93.70 & 95.50 & 60.55 & 77.57 & 80.68 & 52.28 & 68.15 & 74.41 & 40.00 & 55.96 & 64.69 & 47.45 & 96.35 & 96.65 \\ %%
Set-BSR & 50.24 & 66.99 & 72.15 & 81.50 & 94.80 & 96.20 & 61.29 & 78.51 & 81.66 & 52.91 & 68.98 & 75.31 & 40.48 & 56.64 & 65.48 & 48.00 & 96.60 & 97.50 \\ %%
MDL & 49.80 & 66.34 & 71.68 & 81.10 & 94.30 & 95.70 & 61.18 & 78.03 & 81.41 & 53.17 & 68.66 & 75.14 & 39.78 & 56.48 & 65.34 & 48.55 & 96.80 & 97.55 \\
MDR & 49.63 & 66.31 & 71.54 & 79.80 & 94.10 & 95.30 & 61.07 & 78.11 & 81.23 & 53.23 & 68.61 & 75.09 & 40.12 & 56.12 & 65.31 & 48.50 & 96.65 & 97.40 \\
ConE & 50.11 & 66.75 & 71.91 & 81.30 & 94.50 & 95.90 & 61.31 & 78.35 & 81.75 & 53.30 & 68.74 & 75.28 & 40.29 & 56.93 & 65.52 & 48.65 & 96.95 & 97.70 \\
TopicK & \textbf{51.84}$^*$ & \textbf{67.32}$^+$ & \textbf{72.97}$^*$ & \textbf{81.80}$^+$ & \textbf{94.90} & \textbf{96.40} & \textbf{62.04}$^*$ & \textbf{79.63}$^*$ & \textbf{82.68}$^*$ & \textbf{53.46}$^+$ & \textbf{69.35}$^*$ & \textbf{75.59}$^+$ & \textbf{41.30}$^*$ & \textbf{57.85}$^*$ & \textbf{66.34}$^*$ & \textbf{49.85}$^*$ & \textbf{97.15} & \textbf{98.30}$^+$ \\
\bottomrule
\end{tabular}}
\caption{Performance (accuracy) of ICL with different demonstration selection strategies. ``-B'' indicates the model size, and the best result in each column is highlighted in \textbf{bold}. $*$ and $+$ indicate $p \leq $ 0.01 and $p \leq $ 0.05 for the paired t-test with the best competitor.}
\label{tab:main-results}
\end{table*}

\paragraph{Datasets}
We evaluate our method on 6 datasets spanning a variety of domains.
For general-domain tasks, we use \textbf{CommonsenseQA} \citep{talmor-etal-2019-commonsenseqa} and \textbf{SciQ} \citep{SciQ} for natural language understanding, as well as \textbf{QNLI} \citep{qnli} and \textbf{MNLI} \citep{williams-etal-2018-broad} for natural language inference.
To assess question-answering performance in specialized domains, we include \textbf{MedMCQA} \citep{medmcqa} from the medical domain and \textbf{Law} \citep{law} from the legal domain.
Each dataset has a demonstration pool of input-output pairs, and we evaluate the accuracy on the test set with \textit{three} different random seeds.
If the test set is private, we report the results on the validation set as done in \citet{peng-etal-2024-revisiting}.

\paragraph{Baselines}
We compare \textbf{TopicK} (ours) with various conventional and state-of-the-art approaches.
Specifically, we adopt two basic methods:
\begin{itemize}[leftmargin=10pt, itemsep=0pt]
    \item \textbf{Zero} uses no demonstration and serves as a zero-shot baseline.
    \item \textbf{Rand} randomly selects demonstrations for each test example.
\end{itemize}
and four similarity-based approaches:
\begin{itemize}[leftmargin=10pt, itemsep=0pt]
    \item \textbf{BM25} \citep{bm25} selects demonstrations based on lexical overlap.
    \item \textbf{TopK} \citep{liu-etal-2022-makes} selects the $K$-nearest-neighbors using dense retriever embeddings.
    \item \textbf{CEIL} \citep{ceil} adopts DPP \citep{dpp} to enhance diversity. For a fair comparison, we exclude the retriever fine-tuning and apply only the DPP-based inference
    \item \textbf{Set-BSR} \citep{gupta-etal-2023-coverage} selects demonstrations based on BERTScore-Recall (BSR) \citep{bertscore}, to cover the tokens in the test input.
\end{itemize}
and three uncertainty-based approaches:
\begin{itemize}[leftmargin=10pt, itemsep=0pt]
    \item \textbf{MDL} \citep{iter-etal-2023-context} selects demonstrations that minimize predictive uncertainty.
    \item \textbf{MDR} \citep{wang-etal-2024-mdr} selects demonstrations where the model exhibits minimum predictive error.
    \item \textbf{ConE} \citep{peng-etal-2024-revisiting} selects demonstrations that minimize uncertainty on test input.
\end{itemize}
% We select topic-coherent demonstrations based on unsupervised topic modeling to enhance both relevance and diversity, while keeping the language model and retriever frozen.
For all methods compared, we adopt the \textbf{8-shot} setting, following ConE.
We would like to note that all baselines, like ours, freeze both the retriever and the LLMs. %, without any additional fine-tuning
For a fair comparison, we exclude methods utilizing retriever update for selecting prompts \citep{rubin-etal-2022-learning} or demonstrations \citep{chen-etal-2024-learning-retrieve, wang-etal-2024-learning, MOD}.

\paragraph{Implementation Details}
Our evaluation setup, including prompt templates and inference procedures, is based on the OpenICL library~\citep{wu-etal-2023-openicl}.
For the embeddings, we use \texttt{all-mpnet-base-v2} (SBERT) \citep{sbert}, which has shown strong retrieval performance in ConE \citep{peng-etal-2024-revisiting}.
For similarity-based baselines, we utilize the FAISS library~\citep{faiss} to perform efficient nearest-neighbor search.
For uncertainty-based methods, we retrieve 30 candidate demonstrations with TopK to narrow the search space, following their implementations.
All baselines are implemented using publicly available author code, and we strictly follow the documented configurations and hyperparameters.

\begin{figure}[t]
  \centering
  \includegraphics[width=1\linewidth]{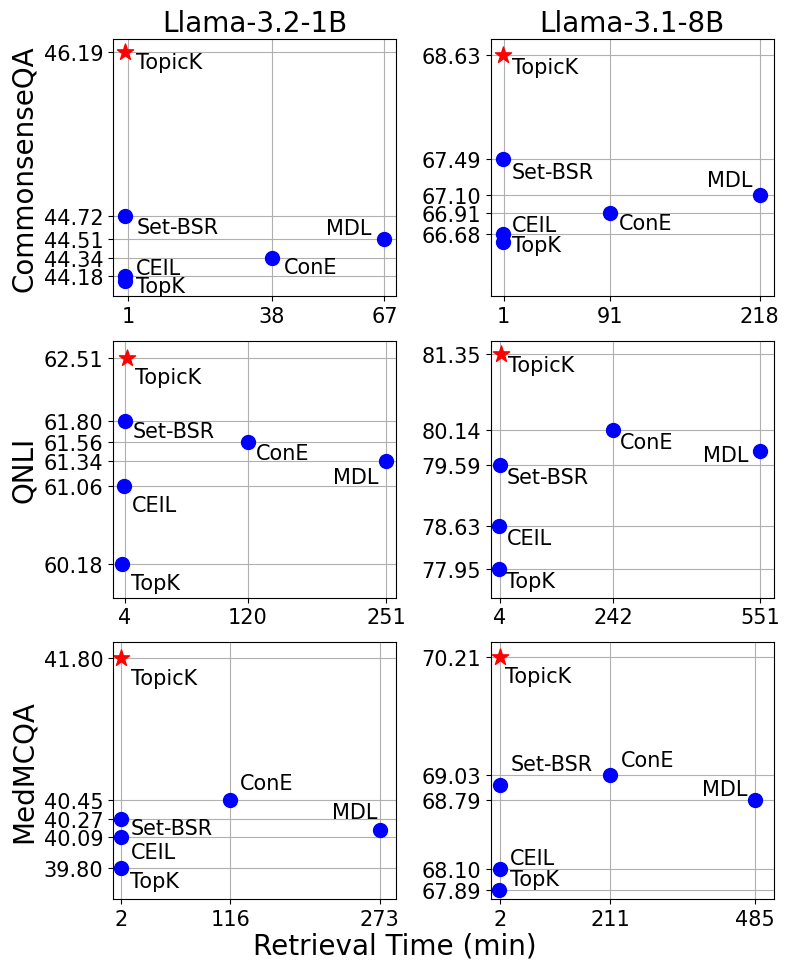}
  \caption{Time-accuracy trade-off. The y-axis represents ICL accuracy, and the x-axis indicates the time elapsed for retrieval with a single A100 GPU.}
  \vspace{-0.3cm} %%%
\label{fig:time}
\end{figure}

\subsection{Main Results}
Table~\ref{tab:main-results} shows the ICL performance with different demonstration selections.
We first observe that different model families possess varying levels of domain knowledge \cite{kweon2025uncertainty}.
For instance, Llama-3.2 models outperform Qwen-2.5 models on MedMCQA but underperform on CommonsenseQA.
This supports our motivation to examine the topical knowledge of each model before retrieving demonstrations.
% By doing so, TopicK provides context that accounts for both the knowledge required by the test input and the knowledge already embedded in the model.
Additionally, we find that Rand occasionally performs worse than Zero, highlighting the importance of appropriate demonstration selection.

Similarity-based approaches (e.g., Set-BSR) generally perform well on general-domain tasks such as CommonsenseQA and SciQ.
In contrast, uncertainty-based methods (e.g., ConE) perform better in specialized domains like MedMCQA and Law.
This is because similarity-based methods rely on surface-level relevance, which is often sufficient for general-domain tasks, whereas uncertainty-based methods incorporate the model’s internal knowledge and uncertainty, making them more effective in handling complex or domain-specific reasoning required in specialized tasks.

TopicK consistently outperforms the state-of-the-art similarity-based (Set-BSR) and uncertainty-based (ConE) methods by integrating semantic similarity with topic coverage.
Across all datasets, TopicK achieves relative improvements of 1.59\% over Set-BSR and ConE.
Notably, TopicK yields larger improvement in specialized domains (MedMCQA, Law) by up to 6.38\% over ConE with Llama-3.2-1B.
This indicates that TopicK selects demonstrations that comprehensively cover the topics in the test input, enabling better leveraging of domain-specific knowledge for unseen tasks.

% \begin{table}[t]
% \centering
% \resizebox{\columnwidth}{!}{
% \begin{tabular}{c|c|ccc}
% \toprule
% \textbf{Model} & \textbf{Method} & \textbf{Common} & \textbf{QNLI} & \textbf{MedMCQA} \\
% \toprule
% \multirow{6}{*}{\shortstack{Gemini-2.0\\-Flash-Lite}}
%   & Zero & 62.33 & 74.17 & 70.31\\
%   & Rand & 65.10 & 76.66 & 71.02\\
%   & TopK & 67.98 & 77.54 & 74.29\\
%   & CEIL & 68.06 & 79.97 & 75.04\\
%   & Set-BSR & 68.23 & 80.36 & 75.47\\
%   & TopicK & \textbf{69.37} & \textbf{84.20} & \textbf{78.59}\\
% \midrule
% \multirow{6}{*}{\shortstack{Claude-3.0\\-Haiku}}
%   & Zero & 57.00 & 72.34 & 53.73\\
%   & Rand & 58.97 & 74.83 & 59.48\\
%   & TopK & 63.64 & 75.71 & 66.80\\
%   & CEIL & 64.78 & 78.14 & 67.65\\
%   & Set-BSR & 65.02 & 78.36 & 67.81\\
%   & TopicK & \textbf{67.40} & \textbf{82.37} & \textbf{69.11}\\
% \bottomrule
% \end{tabular}}
% \caption{Performance of 5-shot ICL with closed-source LLMs. ``Common'' represents CommonsenseQA.}
% \label{tab:closed}
% \end{table}

\begin{table}[t]
\centering
\resizebox{\columnwidth}{!}{
\begin{tabular}{c|c|ccc}
\toprule
\textbf{Model} & \textbf{Method} & \textbf{Common} & \textbf{QNLI} & \textbf{MedMCQA} \\
\toprule
\multirow{6}{*}{\shortstack{Gemini-2.0\\-Flash-Lite}}
  & Zero & 62.33 & 74.17 & 70.31\\
  & Rand & 65.10 & 76.66 & 71.02\\
  & TopK & 67.98 & 77.54 & 74.29\\
  & CEIL & 68.06 & 79.97 & 75.04\\
  & Set-BSR & 68.23 & 80.36 & 75.47\\
  & TopicK & \textbf{69.37} & \textbf{84.20} & \textbf{78.59}\\
\midrule
\multirow{6}{*}{\shortstack{Claude-3.0\\-Haiku}}
  & Zero & 57.00 & 72.34 & 53.73\\
  & Rand & 58.97 & 74.83 & 59.48\\
  & TopK & 63.64 & 75.71 & 66.80\\
  & CEIL & 64.78 & 78.14 & 67.65\\
  & Set-BSR & 65.02 & 78.36 & 67.81\\
  & TopicK & \textbf{67.40} & \textbf{82.37} & \textbf{69.11}\\
\midrule
\multirow{6}{*}{\shortstack{GPT-4o\\-mini}}
  & Zero & 65.52 & 76.36 & 61.68\\
  & Rand & 66.01 & 77.89 & 64.24\\
  & TopK & 69.21 & 82.47 & 70.34\\
  & CEIL & 69.75 & 85.02 & 71.16\\
  & Set-BSR & 70.18 & 85.53 & 71.72\\
  & TopicK & \textbf{70.88} & \textbf{86.54} & \textbf{72.44}\\
\bottomrule
\end{tabular}}
\caption{Performance of 5-shot ICL with closed-source LLMs. ``Common'' represents CommonsenseQA.}
\label{tab:closed}
\end{table}

\begin{figure*}[ht!]
  \centering
  \includegraphics[width=1\linewidth]{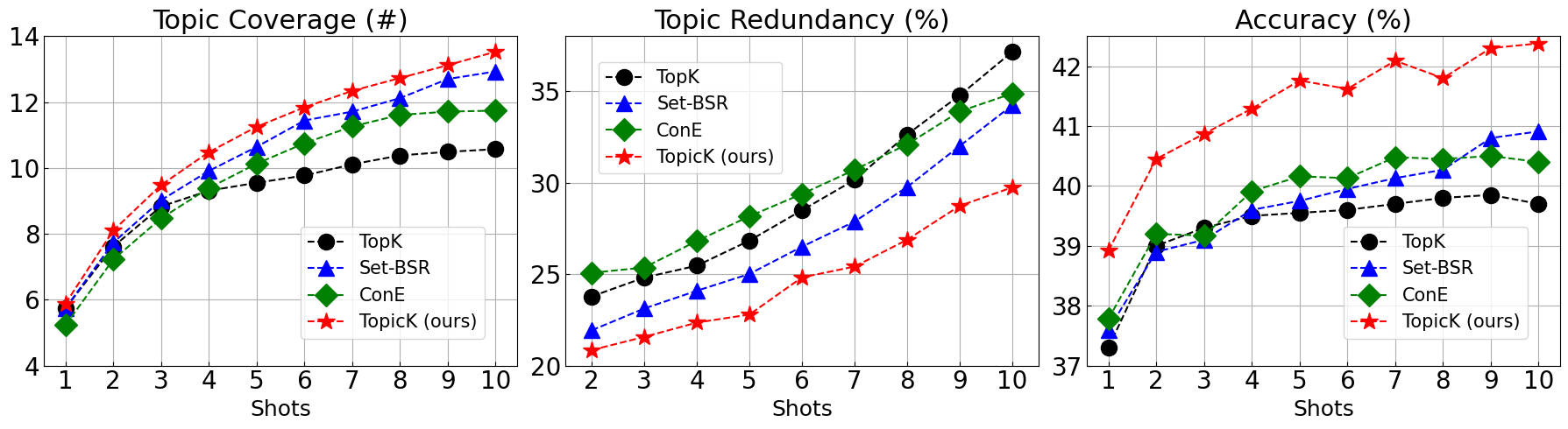}
  \caption {Analysis on topic coverage with Llama-3.2-1B and MedMCQA.}
\label{fig:shots}
\end{figure*}

\subsection{Time-Accuracy Trade-off}
Figure~\ref{fig:time} illustrates the time-accuracy trade-off of Llama-3.2-1B and Llama-3.1-8B across three datasets.
% Here, ``Time'' refers to the retrieval time required for selecting in-context demonstrations, and ``Accuracy'' indicates the ICL performance.
Similarity-based methods (TopK, CEIL, Set-BSR) benefit from efficient retrieval through dual-encoder architectures, offering low latency.
However, their reliance solely on surface-level similarity often leads to suboptimal performance, particularly in specialized domains (i.e., MedMCQA).
On the other hand, uncertainty-based methods (ConE) attain higher accuracy on MedMCQA, by leveraging the LLM itself to evaluate the informativeness of demonstrations.
However, they require separate LLM inference for each demonstration, incurring significant computational overhead at test time; ConE is 37× slower than TopicK on QNLI.

TopicK strikes the best balance between accuracy and efficiency.
TopicK consistently achieves the best performance by comprehensively covering fine-grained topic-level knowledge, while maintaining low retrieval latency.
This advantage arises from its use of a lightweight topic predictor to estimate the knowledge required for each test input.
Importantly, the topic predictor operates independently of the LLM size, making TopicK highly scalable and effective across both small and large LLMs

\subsection{Results with Closed-Source LLMs}
Since TopicK estimates the topical knowledge via a topic predictor and zero-shot accuracy, it can be applied to closed-source LLMs as well. %% 
We note that uncertainty-based methods (MDL, MDR, ConE) rely on generation probabilities, and therefore, are generally incompatible with closed-source LLMs \cite{gemini, claude3, gpt4o}.\footnote{As of submission, logprobs for the first two closed-source LLMs in Table~\ref{tab:closed} are unavailable.}
Table~\ref{tab:closed} presents the 5-shot ICL performance of two closed-source LLMs.
TopicK consistently outperforms all baselines across all tasks and models, demonstrating its effectiveness and generality on restricted APIs.

\subsection{Topic Coverage Analysis}
Figure~\ref{fig:shots} presents an in-depth analysis of demonstration diversity as the number of shots increases from $K=1$ to $10$.
For this analysis, we introduce two metrics:
\begin{itemize}[leftmargin=10pt, itemsep=0pt]
\item \textbf{Topic Coverage}: the number of topics covered by demonstrations (${d_1, \dots, d_K}$), among the top-20 required topics in the test input.
\item \textbf{Topic Redundancy}: the proportion of topics covered by the current demonstration ($d_K$) that have already been covered by previous demonstrations (${d_1, \dots, d_{K-1}}$).
\end{itemize}
\noindent We observe that TopK suffers from low topic coverage and high redundancy due to its reliance on similarity ranking without diversity control.
Set-BSR improves diversity by adopting setwise BERTScore, but remains limited by surface-level embedding similarity and implicit token-level coverage.
ConE, despite considering model uncertainty, shows high redundancy and low coverage as it evaluates each demonstration independently.
In contrast, TopicK explicitly targets fine-grained topic coverage, achieving the highest coverage and lowest redundancy.
This demonstrates its effectiveness in retrieving demonstrations that are not only relevant and informative but also comprehensively cover a broader range of necessary topics, enhancing overall ICL performance.

\begin{table}[t]
\centering
\resizebox{\columnwidth}{!}{
\begin{tabular}{lccc}
\toprule
 & \textbf{Common} & \textbf{QNLI} & \textbf{MedMCQA} \\
\toprule
TopicK & \textbf{46.19} & \textbf{62.51} & \textbf{41.80} \\
w/o Core Topic (\S\ref{sec3.1.1}) & 44.72 & 62.03 & 41.17 \\
w/o Soft Label (\S\ref{sec3.1.2})& 45.21 & 62.38 & 41.56 \\
w/o Topical Knowledge (\S\ref{sec3.2.1})& 44.86 & 60.68 & 40.55 \\			
w/o Cumulative Coverage (\S\ref{sec3.2.2}) & 44.41 & 61.47 & 40.12 \\
\bottomrule
\end{tabular}}
\caption{Ablation study of TopicK with Llama-3.2-1B. ``Common'' represents CommonsenseQA.}
\label{tab:ablation}
\end{table}

\begin{table*}[ht]
\centering
\resizebox{\textwidth}{!}{
\begin{tabular}{p{0.49\textwidth}|p{0.49\textwidth}}
\toprule
\multirow{2}{*}{
  \parbox{0.48\textwidth} {
    \textbf{Test input} (Zero-shot PPL: 2.872) \newline
    Question: Non-human organisms that mainly consume plants/other primary producers are known as what? \newline
    (A) Amphibian \newline (B) Omnivore \newline (C) Herbivore \newline (D) Carnivore}
}
& \textbf{Inferred required topics} \newline
\textcolor{blue}{\textbf{carnivore}} (0.91), \textcolor{red}{\textbf{omnivore}} (0.90), \textcolor{Green}{\textbf{herbivore}} (0.87), 
\textcolor{Green}{\textbf{plant}} (0.34), \textcolor{blue}{\textbf{ecosystem}} (0.28), \textcolor{blue}{\textbf{food chain}} (0.23), 
\textcolor{blue}{\textbf{animal}} (0.18), \textcolor{Green}{\textbf{food web}} (0.09), 
\textcolor{red}{\textbf{insectivore}} (0.07), \textcolor{Green}{\textbf{vegetarian}} (0.06), 
\textcolor{blue}{\textbf{organism}} (0.05) \\
\cmidrule{2-2}
& \textbf{Topical knowledge of LLM} \newline
carnivore (0.72), omnivore (0.85), herbivore (0.75), plant (0.77), ecosystem (0.69), food chain (0.74), 
animal (0.78), food web (0.89), insectivore (0.73), vegetarian (0.76), organism (0.73) \\
\midrule
\midrule
\textbf{Top-1 demonstration} (1-shot PPL: 2.152)\newline
Question: What do you call an animal that feeds on other animals? (A) Carnivore (B) Omnivore (C) Polyvore (D) Herbivore \newline Answer: (A) & 
\textbf{Inferred covered topics} \newline
\textcolor{blue}{\textbf{carnivore}} (0.87), \textcolor{blue}{\textbf{ecosystem}} (0.32), \textcolor{blue}{\textbf{animal}} (0.19), \textcolor{blue}{\textbf{food chain}} (0.19), polyvore (0.13), \textcolor{blue}{\textbf{organism}} (0.11), omnivore (0.08), herbivore (0.07) \\
\midrule
\textbf{Top-2 demonstration} (2-shot PPL: 1.630) \newline
Question: Herbivores are heterotrophs that eat only or mainly what? (A) Plants (B) Animals (C) Fish (D) Decayed matter \newline Answer: (A)  &
\textbf{Inferred covered topics} \newline 
\textcolor{Green}{\textbf{herbivore}} (0.96), heterophile (0.51), \textcolor{Green}{\textbf{plant}} (0.27), xerophyte (0.23), \textcolor{Green}{\textbf{vegetarian}} (0.21), decayed matter (0.19), rotifer (0.15), \textcolor{Green}{\textbf{food web}} (0.08), eutroph (0.07), moss (0.06)\\
\midrule
\textbf{Top-3 demonstration} (3-shot PPL: 1.369) \newline
Question: Omnivores are animals that eat both plant- and? (A) Biofuel (B) Liquid diets (C) Recycled food (D) Animal-derived food \newline Answer: (D) & 
\textbf{Inferred covered topics} \newline
\textcolor{red}{\textbf{omnivore}} (0.90), liquid diet (0.33), recycled food (0.17), animal (0.13), biofuel (0.11), carnivore (0.07), plant (0.07), \textcolor{red}{\textbf{insectivore}} (0.06), food chain (0.03), vegetarian (0.03) \\
\bottomrule
\end{tabular}}
\caption{Detailed case study on SciQ dataset and Llama-3.2-1B (extended from Figure~\ref{fig:intro}). PPL denotes the perplexity (lower is better) for the correct answer ((C) Herbivore). Scores for inferred topics represent importance according to the topic predictor. i.e., $\hat{\mathbf{t}}_{x,t}$ and $\hat{\mathbf{t}}_{d,t}$.}
\vspace{-0.1cm}
\label{tab:casestudy}
\end{table*}

\subsection{Ablation Study}
Table~\ref{tab:ablation} presents an ablation study of TopicK with three variations:
\begin{itemize}[leftmargin=10pt, itemsep=0pt]
    \item \textbf{``w/o Core Topic''} replaces the LLM-matched core topic set with a BM25-generated candidate topic set.
    \item \textbf{``w/o Soft Label''} trains the topic predictor using a binary vector, rather than the distinctiveness-aware soft label (Eq.~\ref{eq:softlabel}).
    \item \textbf{``w/o Cumulative Coverage''} omits the update of covered topics (Eq.~\ref{eq:diversity}), selecting demonstrations independently.
\end{itemize}
\noindent Removing any component reduces performance, confirming their utility.
Removing core topic matching leads to performance degradation across all datasets, confirming the importance of aligning demonstrations with the central topic of the test input.
Eliminating distinctiveness-aware labeling slightly reduces accuracy, suggesting that filtering out popular topics improves selection precision.
Lastly, the removal of the cumulative topic coverage consistently causes the largest degradation, especially on MedMCQA (-4.02\%), indicating that capturing a wide range of subtopics is crucial for complex knowledge-intensive tasks.

\subsection{Caset Study}
Table~\ref{tab:casestudy} presents a case study with TopicK, extending Figure~\ref{fig:intro}. 
First, we observe the topic predictor generalizes to unseen inputs and enriches topic distributions by inferring implicitly related concepts (e.g., \texttt{ecosystem}, \texttt{food chain}). 
Second, TopicK retrieves diverse, non-redundant demonstrations that ensure broad topic coverage. 
Third, it incorporates topical knowledge, selecting demonstrations where the model shows weaker understanding. 
For instance, TopicK prefers \textcolor{Green}{\textbf{herbivore}} (0.87) over \textcolor{red}{\textbf{omnivore}} (0.90), considering the model’s lower topical knowledge on herbivore (0.75 vs. 0.85).

% Table~\ref{tab:casestudy} presents a detailed case study with TopicK, extending from Figure~\ref{fig:intro}.
% First, we observe that the topic predictor not only generalizes to the unseen test input but also enriches the topic distribution by inferring semantically related concepts.
% For example, TopicK identifies relevant topics such as \texttt{ecosystem} and \texttt{food chain}, which are not explicitly mentioned in the input question but enhance the model’s understanding.

% Second, TopicK retrieves a diverse and relevant set of demonstrations that comprehensively cover the required topics in the test input.
% By considering cumulative topic coverage, TopicK avoids retrieving redundant demonstrations and prioritizes those that address uncovered but important topics.

% Third, TopicK incorporates the model's understanding of each topic via topical knowledge.
% For instance, although \textcolor{red}{\textbf{omnivore}} (0.90) has a higher importance than \textcolor{Green}{\textbf{herbivore}} (0.87), the model exhibits weaker topical knowledge for herbivore.
% As a result, TopicK selects the herbivore-related demonstration (2-shot PPL: 1.630) over the omnivore-related one (2-shot PPL: 1.820, if the Top-2 and Top-3 demonstrations are swapped).

\section{Conclusions}
We argue that an effective set of demonstrations should provide comprehensive coverage of fine-grained aspects (i.e., topics) required by the test input and language models.
To this end, we propose TopicK, which identifies the required topics in the test input and retrieves demonstrations that maximize the cumulative topic coverage.
By assessing the model’s informational needs through topic-level signals, TopicK relies solely on a lightweight topic predictor and avoids any LLM inference at test time.
Extensive experiments across diverse domains and both open- and closed-source LLMs demonstrate that TopicK consistently outperforms state-of-the-art methods.

\section*{Acknowledgments}
This work was supported by Samsung Research Funding \& Incubation Center of Samsung Electronics under Project Number SRFC-IT2402-05 (South Korea), and by Molecule Maker Lab Institute: An AI Research Institutes program supported by NSF under Award No. 2019897 (United States).

\section*{Limitations}
\paragraph{Model scale} Due to computational constraints, we evaluate TopicK on open-source LLMs ranging from 0.5B to 8B parameters.
It would be valuable to scale our experiments to larger models such as Llama-3.3-70B.
% Instead, we validate TopicK on large-scale closed-source models, including Gemini-2.0-Flash-Lite and Claude-3.0-Haiku.

\paragraph{Flat topic set} In this work, we construct a flat topic set and devise a flat topic predictor.
Exploring hierarchical topic structures (e.g., topical taxonomy) remains a promising direction for future work, potentially enabling a richer understanding of topic coverage.

% This work was also supported by Basic Science Research Program through the NRF funded by the Ministry of Education (NRF-2021R1A6A1A03045425) and ICT Creative Consilience Program through the IITP grant funded by the MSIT (IITP-2025-RS-2020-II201819).

\bibliography{anthology,custom}
% \bibliography{custom}

\clearpage  % Start a new page

\appendix
\section{Implementation Details for TopicK}
Our source code, including the core topic set for each demonstration, is available at \textcolor{blue}{\url{https://github.com/WonbinKweon/TopicK_EMNLP2025}}

\paragraph{Topic Mining}
We employ two topic mining tools: SeedTopicMine \citep{seedtopicmine} for extracting single-word topics and AutoPhrase \citep{autophrase} for multi-word phrases.
We then merge the outputs to construct the topic set $\mathcal{T}$ for each dataset.

\paragraph{Core Topic Matching with GPT-4o}
We prompt GPT-4o to select the core topics from the candidate topic set $\mathcal{T}'_d$ and identify any missing but relevant topics, as follows:
% \begin{figure}[h]
%   \centering
%   \includegraphics[width=1\linewidth]{Figures/prompt.jpg}
%   \caption{Prompt for core topic matching.}
%   \vspace{-0.3cm} %%%
% \label{fig:prompt}
% \end{figure}
\begin{table}[h]
\centering
\resizebox{1.0\columnwidth}{!}{
\begin{tabular}{|p{\linewidth}|}
\hline
You will receive a question-answer demonstration along with a candidate topic set. Your task is to output relevant topics of the demonstration. You may choose topics from the candidate topic set, or you can create new relevant topics. You must provide at least five topics. Do not include any explanation or numbers. Please just output the list of relevant topics, separated by commas. Demonstration:~$\{d\}$, Candidate~topic~set:~$\{\mathcal{T}'_d\}$ \\
\hline     
\end{tabular}}
\vspace{-0.1cm}
\label{tab:prompt}
\end{table}

\noindent This process yields the finalized core topic set $\mathcal{T}_d \subset \mathcal{T}$ for each demonstration $d$.

\paragraph{Topic Predictor}
In this work, we employ a three-layer MLP $\hat{\mathbf{t}}_d = f(\mathbf{e}_d)$ for the topic predictor.
The input embedding $\mathbf{e}_d \in \mathbb{R}^{768}$ is extracted using the \texttt{all-mpnet-base-v2} model \citep{sbert}.
The last classification layer is initialized with the embeddings of topic names.
The model is optimized using the distinctiveness-aware soft label (Eq.\ref{eq:softlabel}) and binary cross-entropy (Eq.\ref{eq:BCE}).

\section{Experiment Details}\label{sec:expdetail}

\paragraph{Datasets}
Table~\ref{tab:dataset_stats} shows the statistics of each dataset.
``\#Topics'' indicates the number of mined topics from each dataset.
Since Law \citep{law} dataset does not provide an explicit data split, we randomly partition the 10k input-output pairs into training and test sets using an 8:2 ratio.
All datasets are sourced from their official HuggingFace repositories~\citep{huggingface}.

\begin{table}[t]
\centering
\resizebox{\columnwidth}{!}{
\begin{tabular}{cccc}
\toprule
\textbf{Dataset} & \textbf{Data Split} & \textbf{\#Classes} & \textbf{\#Topics} \\
\toprule
CommonsenseQA & 9,741 / 1,221 & 5 & 3,781 \\
SciQ          & 11,679 / 1,000 & 4 & 11,451 \\
QNLI          & 104,743 / 5,463 & 2 & 51,809 \\
MNLI          & 392,702 / 9,815 & 3 & 109,390 \\
MedMCQA       & 120,765 / 2,816 & 4 & 49,925 \\
Law           & 8,000 / 2,000 & 4 & 5,296 \\
\bottomrule
\end{tabular}}
\caption{Dataset statistics.}
\label{tab:dataset_stats}
\end{table}

\begin{table}[t]
\centering
\resizebox{\columnwidth}{!}{
\begin{tabular}{ccc}
\toprule
\textbf{Task} & \textbf{Prompt} & \textbf{Class} \\
\toprule
\multirow{5}{*}{CommonsenseQA} 
& Question: $<x>$ Answer: <A> & A \\
& Question: $<x>$ Answer: <B> & B \\
& Question: $<x>$ Answer: <C> & C \\
& Question: $<x>$ Answer: <D> & D \\
& Question: $<x>$ Answer: <E> & E \\
\midrule
\multirow{3}{*}{MNLI}
& $<x_1>$ Can we know $<x_2>$? Yes. & Entailment \\
& $<x_1>$ Can we know $<x_2>$? Maybe. & Neutral \\
& $<x_1>$ Can we know $<x_2>$? No. & Contradiction \\
\midrule
\multirow{2}{*}{QNLI}
& $<x_1>$ Can we know $<x_2>$? Yes. & Entailment \\
& $<x_1>$ Can we know $<x_2>$? No. & Contradiction \\
\midrule
\multirow{4}{*}{\shortstack{SciQ\\MedMCQA\\Law}}
& Question: $<x>$ Answer: <A> & A \\
& Question: $<x>$ Answer: <B> & B \\
& Question: $<x>$ Answer: <C> & C \\
& Question: $<x>$ Answer: <D> & D \\
\bottomrule
\end{tabular}}
\caption{Templates of tasks. $<x>$ is a placeholder for test inputs.}
\vspace{-0.1cm}
\label{tab:templates}
\end{table}

\paragraph{Templates}
% OpenICL
% PPL inference
We adopt the OpenICL library~\citep{wu-etal-2023-openicl} for the prompt templates and inference procedures.
Table~\ref{tab:templates} shows the templates in OpenICL for datasets in the experiment.
For a stable evaluation, following ConE \citep{peng-etal-2024-revisiting}, we adopt the perplexity-based inference in OpenICL.

\paragraph{Hyperparameters}
All hyperparameters of TopicK and baselines are selected with a grid search on the validation set.
If the test set is private and the validation set is used for evaluation, we reserve 10\% of the training set as a held-out validation set.
For CEIL, the scale factor $\lambda$ for the DPP-based inference is selected from [0, 0.5].
For uncertainty-based methods (MDL, MDR, ConE), we retrieve 30 candidate demonstrations with TopK to narrow the search space, following their implementations.
For MDL, the select time is set to 10 to constrain the time limit.
For MDR, the coefficient $C$ is selected from $[0, 1]$.
For TopicK, the learning rate of the topic predictor is selected from $[1\text{e}{-5}, 1\text{e}{-4}]$.
For the final relevance score \hyperref[eq:lambda]{$r(x,d) + \lambda \cdot \cos(\mathbf{e}_x, \mathbf{e}_d)$}, $\lambda$ is selected from $[0.1, 1]$ and z-score normalization is applied for $r(x,d)$ and $\cos(\mathbf{e}_x, \mathbf{e}_d)$ to ensure their scales are matched.

\paragraph{Resources}
For open-source LLMs (i.e., Llama3.2 and Qwen2.5 families), all experiments were conducted on a single NVIDIA A100 80GB GPU with an AMD EPYC™ 7513 2.60GHz CPU.  
For closed-source LLMs (i.e., Gemini-2.0-Flash-Lite and Claude-3.0-Haiku, GPT-4o-mini), all experiments were performed via their respective APIs, subject to usage-based pricing.

% \section{Case Study}\label{sec:casestudy}

\end{document}